\documentclass[letterpaper, 10 pt, conference]{ieeeconf}

\IEEEoverridecommandlockouts
\usepackage[utf8]{inputenc}
\usepackage[T1]{fontenc}
\usepackage{booktabs}
\usepackage{arydshln}
\usepackage[flushleft]{threeparttable}
\usepackage{amsmath}
\usepackage{amsfonts}
\usepackage{microtype}
\usepackage{graphicx}
\usepackage{capt-of}
\usepackage{tikz}
\usetikzlibrary{arrows.meta,positioning,fit,calc}
\usepackage{xcolor}
\usepackage{url}
\usepackage{pifont}
\usepackage{hyperref}

\newcommand{\cmark}{\textcolor{black}{\ding{51}}}
\newcommand{\xmark}{\textcolor{black!45}{\ding{55}}}

\graphicspath{{figures/}}

\title{\LARGE \bf
STRIDER: Stepping-Enabled Multi-Gait Hierarchical 3D Loco-Manipulation Framework for Humanoid Robots
\thanks{Video: \url{https://youtu.be/gf5RWjCZXtA}}
}

\author{\authorblockN{Yuanzhuo Li,
Wen Zhao,
Zhe Yong,
Xiang Meng,
Gang Han,\\
Hengle Ren,
Xiaoyang Zheng,
Zhen Wang,
Yijie Guo\authorrefmark{1}}\\
\authorblockA{X-Humanoid, Humanoid Innovation Department}
\authorblockA{\authorrefmark{1} Corresponding author: \texttt{jack.guo@x-humanoid.com}}}

\IEEEaftertitletext{%
  \vspace{0.5\baselineskip}%
  \centering
  \includegraphics[width=0.8\textwidth, height=0.3\textheight, keepaspectratio]{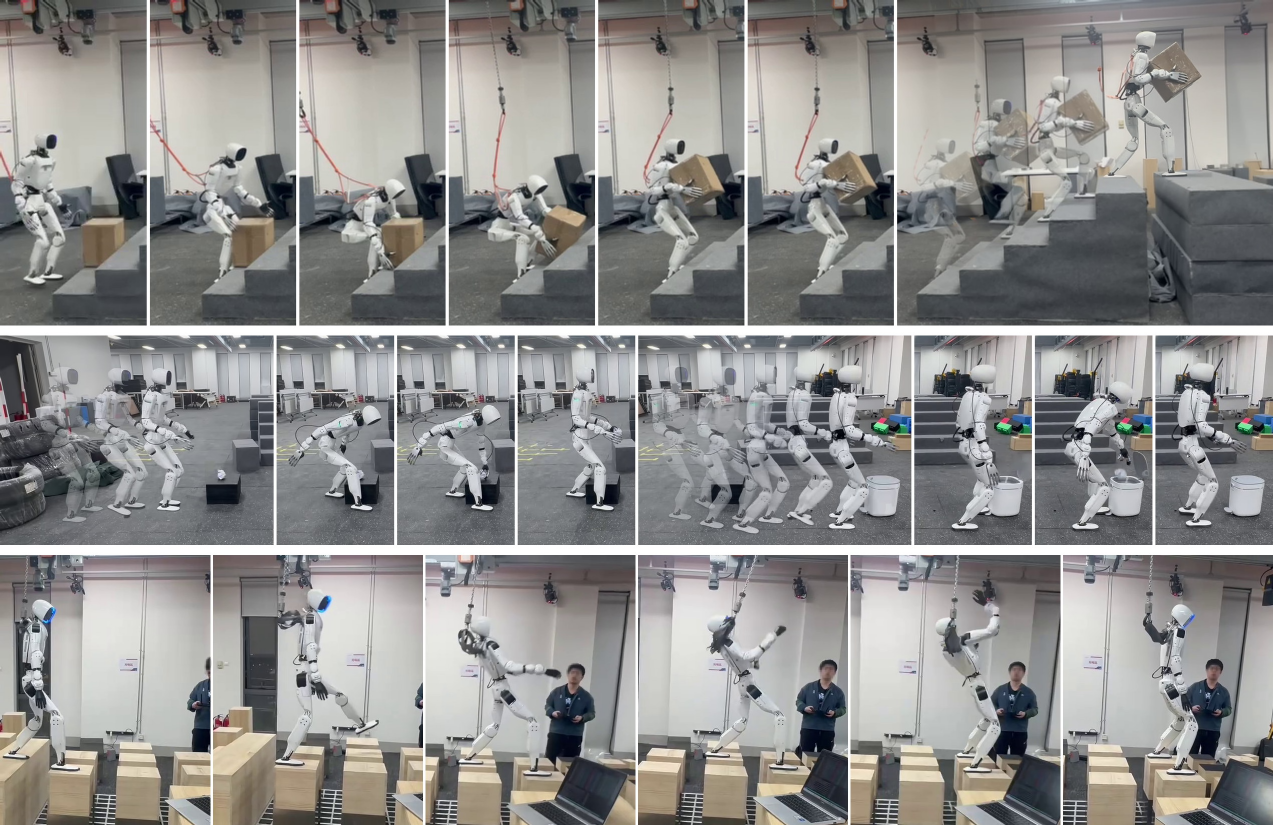}%
  \captionof{figure}{Multi-gait loco-manipulation tasks performed by STRIDER on
  the TianGong Omni humanoid. Top: the robot adjusts its footholds, picks up a
  box, climbs stairs, and crosses a gap using foothold tracking. Middle: it
  uses velocity tracking to approach the trash, adjusts its footholds to pick
  it up, walks to the bin, and uses foothold tracking to step on the pedal and
  deposit the trash. Bottom: it traverses stepping stones using foothold
  tracking while making large arm swings and posture adjustments that mimic
  bullet-dodging motions. A video of the hardware experiments is available at
  \url{https://youtu.be/gf5RWjCZXtA}.}%
  \label{fig:all_tasks}%
  \vspace{0.5\baselineskip}%
}

\begin{document}
\maketitle
\thispagestyle{empty}
\pagestyle{empty}

\begin{abstract}
Humanoid loco-manipulation faces two prominent limitations: controllers using continuous velocity commands cannot precisely regulate individual footholds, while specialized foothold-tracking modules are difficult to integrate with whole-body manipulation. Furthermore, standard action-based imitation distillation primarily transfers expert actions, without explicitly encouraging a shared representation of heterogeneous skills. This paper introduces STRIDER, a hierarchical multi-gait framework to bridge these gaps. The framework integrates terrain-aware 3D stepping logic, Adversarial Motion Priors (AMP)-based natural walking, and Cartesian upper-body control: its stepping expert selects feasible footholds in the stance-foot frame and generates clearance-aware swing trajectories. To fuse distinct walking and stepping experts into one executable student policy, we propose Latent Distillation Proximal Policy Optimization (LD-PPO), a distillation algorithm augmented with teacher-conditioned latent alignment. By jointly optimizing on-policy reinforcement learning, DAgger-based action reconstruction, and latent alignment, LD-PPO transfers expert actions while encouraging a shared skill representation across heterogeneous modes. Simulation and real-robot evaluations on the TianGong Omni humanoid show that LD-PPO outperforms vanilla distillation-PPO in foothold-tracking and posture-tracking accuracy. Deployed on hardware, STRIDER realizes multi-gait loco-manipulation with accurate foothold and end-effector tracking.
\end{abstract}


\section{Introduction}

Reinforcement learning has expanded humanoid-robot control from individual
skills to increasingly versatile whole-body behaviors. Learned locomotion
controllers now negotiate sparse footholds and perform agile parkour
~\cite{wang2025beamdojo,zhuang2024parkour,wu2026perceptive}; manipulation
controllers coordinate the torso and limbs to enlarge the reachable workspace
~\cite{li2025amo}; loco-manipulation systems combine walking with robust
physical interaction~\cite{zhang2026falcon,sun2025ulc}; and versatile
whole-body controllers reproduce task motions or human demonstrations
~\cite{he2024hover,zhao2026halomi}. These developments demonstrate the
potential of reinforcement learning for deploying humanoids in tasks that
require mobility, dexterity, and human-like coordination.

Nevertheless, current loco-manipulation controllers generally drive the lower
body through continuous linear- and angular-velocity commands
~\cite{zhang2026falcon,sun2025ulc,he2024hover}. Such commands are effective for
moving through open space, but do not specify the position, height, or yaw of
an individual foot. Consequently, when a robot approaches an object or moves
over sparse or non-planar support, small stance errors can leave the pelvis in
an unfavorable pose, reduce hand reachability, and compromise stability during
physical interaction. Repeated velocity corrections also provide no guarantee
that the final support configuration is suitable for manipulation. Conversely,
recent learned foothold trackers achieve accurate discrete stepping
~\cite{wang2025beamdojo,montenegro2026mind,crismariu2026march}, but are designed
primarily for locomotion and do not jointly control upper-body manipulation and
torso posture.

Foothold tracking is therefore a necessary complement to velocity-controlled
loco-manipulation. Continuous commands can efficiently bring the robot close
to a task, after which explicit foothold commands can establish a reachable
and stable support pose while coordinating base height and torso orientation.
Although model-based foothold generation and whole-body optimization provide
precise support placement~\cite{kajita2003biped,roux2024mpc}, integrating this
capability with robust learned walking and manipulation in a single deployable
controller remains challenging.

To bridge this gap, we present STRIDER, a hierarchical multi-gait
loco-manipulation framework. The resulting policy supports efficient continuous locomotion, precise terminal
stepping, posture adjustment, and whole-body manipulation. We validate the framework in
simulation and on the TianGong Omni humanoid.

The contributions of this paper are as follows:
\begin{itemize}
  \item STRIDER provides a hierarchical framework that integrates precise 3D stepping, natural walking, posture control, and Cartesian upper-body manipulation.
  \item A terrain-aware stepping controller combines stance-foot-relative commands, foothold filtering, and clearance-aware swing trajectories.
  \item Latent Distillation Proximal Policy Optimization (LD-PPO) transfers both action-level and skill-level information from expert policies to a student policy.
  \item Simulation and TianGong Omni experiments validate foothold-tracking accuracy and multi-gait distillation, as shown in Fig.~\ref{fig:all_tasks}.
\end{itemize}

\section{Related Work}

\subsection{Whole-Body Loco-Manipulation}
Humanoid loco-manipulation extends the workspace of fixed-base manipulators by combining locomotion with physical interaction. Existing systems approach this problem from several directions. FALCON separates upper- and lower-body policies and trains them with a force curriculum for robust interaction~\cite{zhang2026falcon}. AMO combines motion synthesis, trajectory optimization, and reinforcement learning to enlarge the feasible whole-body workspace~\cite{li2025amo}. In contrast, ULC uses one policy with progressive curricula and residual actions to track locomotion and manipulation commands jointly~\cite{sun2025ulc}. HOVER distills multiple command modes into a versatile whole-body controller~\cite{he2024hover}, while HALOMI maps human head--hand demonstrations to feasible motions through a learned behavior manifold~\cite{zhao2026halomi}. Related systems couple walking with expressive or teleoperated upper-body motion~\cite{cheng2024exbody,he2024omnih2o,fu2024humanplus,ben2025homie}.

These methods substantially improve dexterity, robustness, and whole-body coordination. However, many existing loco-manipulation systems rely on continuous walking commands, which provide limited direct control over the final support configuration during manipulation. This can make precise object approach and posture adjustment difficult when the task requires a specific foothold or base pose. STRIDER therefore complements velocity-controlled locomotion with explicit foothold tracking, allowing the robot to approach a task efficiently and then establish a precise support configuration while adjusting base height and torso orientation.

\subsection{Humanoid Foothold Planning}
Explicit foot placement has traditionally relied on model-based planning. ZMP preview control generates dynamically balanced walking patterns~\cite{kajita2003biped}, and linear MPC or DCM-based controllers decide footholds online from a velocity command~\cite{herdt2010online,englsberger2015dcm}. Whole-body MPC can couple online foothold sequencing with full-body dynamics~\cite{roux2024mpc}. Sequential centroidal MPC decouples foothold and contact-wrench optimization to achieve high-frequency perturbation rejection during dynamic biped walking~\cite{meng2026scmpc}. Related work on general legged robots also introduces collision-aware tests that reject footholds without feasible swing trajectories~\cite{ye2026kcfrc}. Learning-based methods improve robustness and transfer. BeamDojo uses staged reinforcement learning and foothold-specific rewards for sparse supports~\cite{wang2025beamdojo}; Mind Your Steps learns a terrain-agnostic 3D foothold tracker that can be paired with different planners~\cite{montenegro2026mind}; and MARCH guides a privileged teacher with model-based references before distilling it into a vision-based student~\cite{crismariu2026march}. 

These works improve stepping accuracy, but remain primarily locomotion systems and do not jointly address foothold placement, torso posture, and upper-body manipulation tasks. STRIDER fills this gap with a terrain-aware stepping expert embedded in a hierarchical loco-manipulation framework.

\begin{figure*}[htbp]
  \centering
  \includegraphics[width=0.8\textwidth,height=0.2\textheight,keepaspectratio]{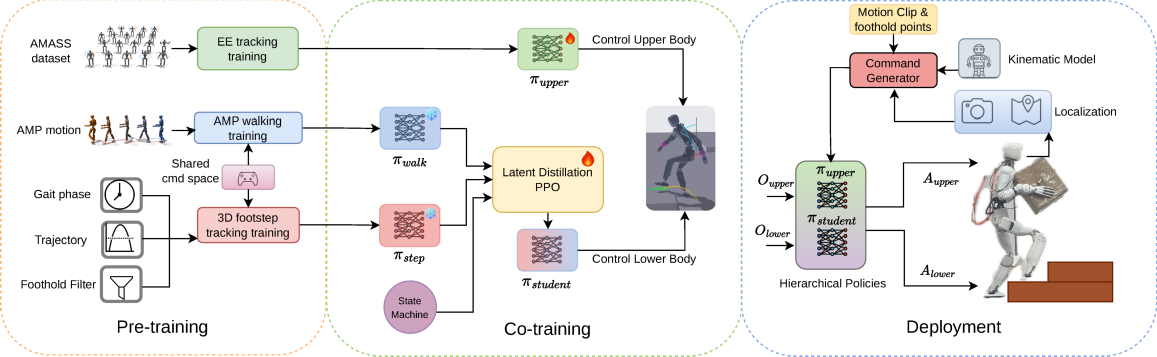}
  \caption{Overview of the hierarchical training and deployment framework.
  Left: the upper-body, AMP-walking, and 3D foothold experts are pretrained
  from motion data and task-specific commands. Center: Latent Distillation PPO
  consolidates the walking and stepping experts into a multi-gait lower-body
  student, which is co-trained with the upper-body policy. Right: a command
  generator converts localization feedback and specified footholds through
  forward kinematics into commanded foothold pose and yaw; motion clips provide
  upper-body targets to $\pi_{\mathrm{upper}}$ and posture commands to
  $\pi_{\mathrm{student}}$. The resulting upper- and lower-body actions jointly
  control the robot.}
  \label{fig:fig1}
\end{figure*}

\subsection{Locomotion with Distillation}
Policy distillation transfers privileged knowledge to deployable policies and can consolidate multiple specialists~\cite{rusu2016policy}. DAgger reduces distribution shift by repeatedly labeling states visited by the student~\cite{ross2011dagger}. Privileged-teacher pipelines similarly distill terrain or extrinsic information into proprioceptive students for sim-to-real locomotion~\cite{lee2020learning,kumar2021rma}. Learn to Teach improves sample efficiency by co-training privileged teachers and students~\cite{wu2025learn}, while Distillation-PPO combines teacher supervision with reinforcement learning under partial observations~\cite{zhang2025dppo}. HOVER distills multiple whole-body command modes~\cite{he2024hover}, and humanoid parkour systems distill privileged motion experts into perceptive multi-skill policies~\cite{zhuang2024parkour,wu2026perceptive}. StyleLoco further uses adversarial distillation to combine agile expert behavior with natural motion priors~\cite{ma2025styleloco}.

Despite these advances, conventional distillation-PPO objectives mainly regress teacher actions pointwise. For heterogeneous walking modes, such supervision leaves the student to infer the shared coordination structure implicitly from action targets. This is particularly challenging for stepping, where small joint-space errors can accumulate into foothold and balance errors. LD-PPO therefore combines DAgger action supervision and on-policy reinforcement learning with teacher-conditioned latent alignment. The key idea is to use the teacher state-action pair to encode the coordination context that produces an expert action, while requiring the deployable student to infer a corresponding latent from observations alone. This separates multi-gait distillation into skill inference and action decoding while retaining action-level supervision.

\section{Methods}

\subsection{Hierarchical Loco-Manipulation Framework}
As shown in Fig.~\ref{fig:fig1}, STRIDER is organized into pre-training,
co-training, and deployment stages. During pre-training, AMASS demonstrations
supervise an end-effector tracking policy $\pi_{\mathrm{upper}}$, while AMP
motions train the velocity-conditioned walking expert $\pi_{\mathrm{walk}}$.
In parallel, the 3D foothold expert $\pi_{\mathrm{step}}$ learns from a shared
command space containing gait phase, swing trajectory, and terrain-filtered
foothold targets. This decomposition allows each expert to specialize in its
own command modality before the policies are coupled.

During co-training, a state machine selects the appropriate lower-body expert
according to the requested gait. Latent Distillation PPO transfers the walking
and stepping experts into a single student policy
$\pi_{\mathrm{student}}$, while $\pi_{\mathrm{upper}}$ controls the upper
body. The upper- and lower-body actions are applied simultaneously in the
simulator, exposing both controllers to the disturbances generated by their
interaction and thereby improving whole-body coordination.

At deployment, motion clips and desired foothold points are converted into
commands using the kinematic model and localization feedback. The hierarchical
controller evaluates $\pi_{\mathrm{upper}}$ and
$\pi_{\mathrm{student}}$ from their respective observations
$O_{\mathrm{upper}}$ and $O_{\mathrm{lower}}$, then combines the resulting
actions $A_{\mathrm{upper}}$ and $A_{\mathrm{lower}}$ to drive the robot.
Consequently, one deployable system supports Cartesian manipulation,
AMP-style walking, and precise 3D stepping without switching the upper-body
controller.

\subsection{3D Foothold Tracking}
To enable discrete foothold tracking across different terrains and provide precise posture control for loco-manipulation, we design a 3D foothold-tracking controller with explicit posture commands.

\subsubsection{Command Space Design}
The 15-D command space comprises planar velocity $(v_x,v_y,v_{\mathrm{yaw}})$, base pose $(h,r,p,\psi)$, and two 4-D foothold targets. Roll and pitch are defined in the gravity-aligned world frame, yaw is defined relative to the pelvis, and height is measured from the supporting foot to the pelvis. Commands $8$--$11$ and $12$--$15$ specify the left- and right-foot targets $(x,y,z,\psi)$ relative to the opposite stance foot.

During standing, velocity and foothold commands are zero, and the current base pose is maintained. During stepping, velocity remains zero, while base-pose and foothold commands are constrained to maintain tracking while preventing leg crossing and excessive foot rotation.

\subsubsection{Observation Space Design}
Both policies share the same observation structure but use different command vectors. The actor receives
$\mathbf{o}_t =
\left[
\boldsymbol{\omega}_b,\,
\mathbf{g}_b,\,
\mathbf{c},\,
\mathbf{q}-\mathbf{q}_0,\,
\dot{\mathbf{q}},\,
\mathbf{a}_{t-1},\,
\sin(2\pi\boldsymbol{\varphi}),\,
\cos(2\pi\boldsymbol{\varphi})
\right]$.
Here, $\boldsymbol{\omega}_b$, $\mathbf{g}_b$, $\mathbf{c}$, $\mathbf{q}-\mathbf{q}_0$, $\dot{\mathbf{q}}$, and $\mathbf{a}_{t-1}$ denote base angular velocity, projected gravity, task command, joint offsets, joint velocities, and the previous action, respectively; $\boldsymbol{\varphi}$ denotes the leg phases. Six frames are stacked.

The asymmetric critic additionally receives privileged information:
\[
\begin{aligned}
\mathbf{o}^{\mathrm{priv}}_t=\bigl[&
\mathbf{o}_t,\mathbf{v}_b,\mathbf{F}_{f,z},
\mathbf{p}_{\mathrm{CoM},xy},\\
&\mathbf{p}_{F,h},\mathbf{F}_{h},
\mathbf{p}_{F,t},\mathbf{F}_{t},\\
&\mathbf{p}_{f},\boldsymbol{\psi}_{f},
\mathbf{h}_{\mathrm{map}}\bigr].
\end{aligned}
\]
The additional terms encode base velocity, vertical foot-contact forces, horizontal center-of-mass position, applied hand and torso forces, foot poses, and an $11\times11$ local terrain-height map.

\subsubsection{Foothold Sampling and Filtering Mechanism}

Selecting accessible and stable footholds is critical for 3D stepping. Unlike planar stepping, terrain height variation can make nominal targets unreachable or unstable. We therefore introduce terrain-aware height generation and local patch filtering to reject unsafe footholds before they are presented to the policy.

\textit{a) Terrain-aware foothold height generation:}
Unlike terrain-independent velocity commands, foothold targets must conform to the local terrain. The terrain height map \(H\) is queried at each sampled horizontal foothold location \(x, y\) to determine its target height \(z\). Near terrain edges, the minimum neighboring height is used to avoid placing the target above the support surface.

\textit{b) Candidate foothold patch evaluation:}
Terrain-aware targets may still be unsafe due to rough surfaces, cliffs, or unreachable elevations. We therefore evaluate each candidate patch by sampling terrain heights along the foot contour and applying three terrain-specific safety criteria.

\begin{figure}[htbp]
  \centering
  \includegraphics[scale=0.07]{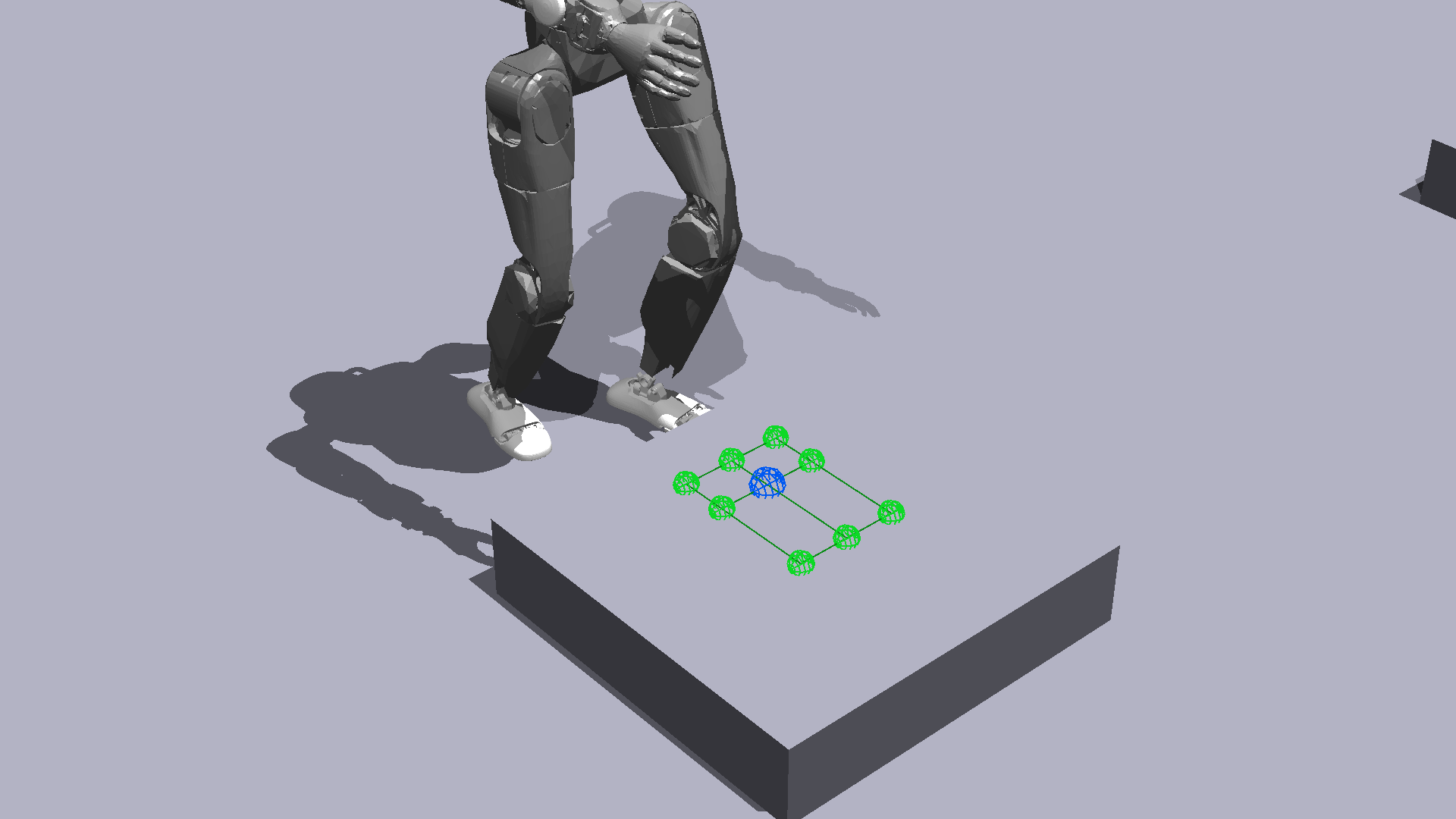}
  \caption{Terrain-filter sampling patch around a candidate foothold. The eight surrounding samples are shown in green, and the blue point denotes the candidate foothold center.}
  \label{fig:foothold-patch-samples}
\end{figure}

Around each candidate foothold, a $3\times 3$ grid of height samples is formed over an inflated foot envelope, as illustrated in Fig.~\ref{fig:foothold-patch-samples}, so that accepted targets remain inside a conservative support region.

We use three criteria to evaluate the local terrain around each candidate foothold and reject unsafe targets, as defined in~\eqref{eq:criterion_level}--\eqref{eq:criterion_reachability}.

\textit{1) Patch sampling:}
Let $\mathbf{p}=(x,y)$ be the horizontal position of a candidate foothold and
$\psi_f$ its yaw command relative to the stance foot. With rear, front, and
lateral sole extents $l_r$, $l_f$, and $w$, and inflation factor $\kappa$, the
half-extents of the sampling envelope are
\begin{equation}
\begin{aligned}
a&=\kappa\bigl(l_r\cos|\psi_f|+w\sin|\psi_f|\bigr),\\
b&=\kappa\bigl(l_f\cos|\psi_f|+w\sin|\psi_f|\bigr),\\
c&=\kappa\bigl(\max(l_f,l_r)\sin|\psi_f|
  +w\cos|\psi_f|\bigr).
\end{aligned}
\label{eq:patch_envelope}
\end{equation}
Using the envelope in~\eqref{eq:patch_envelope}, the nine samples are placed at the offsets $\Delta x\in\{-a,0,b\}$ and
$\Delta y\in\{-c,0,c\}$, indexed in row-major order $j=3i+k+1$ with
$i,k\in\{0,1,2\}$:
\begin{equation}
h_j=H\!\left(x+\Delta x_i,\;y+\Delta y_k\right),\qquad j=1,\dots,9,
\label{eq:patch_heights}
\end{equation}
where $H$ is the conservative height lookup defined above.

\textit{2) Safety evaluation:}

The first criterion $C_1$ requires the patch to lie within a single height band,
so that the whole sole is supported at one level:
\begin{equation}
C_1:\qquad S=\max_{j} h_j-\min_{j} h_j\;\le\;\epsilon.
\label{eq:criterion_level}
\end{equation}
The second criterion $C_2$ accepts surfaces whose height varies approximately
as a plane over the sampled patch. Unlike a one-dimensional traversal of the
grid, we compute finite differences independently along the two horizontal
directions. Let $h_{ik}$ denote the sample at
$(x+\Delta x_i,y+\Delta y_k)$, with $i,k\in\{0,1,2\}$. The directional
height differences are
\begin{equation}
\begin{gathered}
C_2:\quad
d^x_{ik}=h_{i,k+1}-h_{i,k},
\qquad d^y_{ik}=h_{i+1,k}-h_{i,k},
\\[-0.2em]
\max_{i,k\in\{0,1\}}
\left\{
\left|d^x_{i+1,k}-d^x_{i,k}\right|,
\left|d^y_{i,k+1}-d^y_{i,k}\right|
\right\}\leq\epsilon.
\end{gathered}
\label{eq:criterion_slope}
\end{equation}
Here, the first term checks consistency of the slope in the $y$ direction
across adjacent rows, while the second checks consistency of the slope in the
$x$ direction across adjacent columns. The criterion therefore retains
approximately level or steadily inclined patches without depending on the
row-major ordering of the samples.
The third criterion $C_3$ restricts the elevation change to the range the swing
leg can actually reach. Let $h_{\mathrm{stance}}$ denote the terrain height
under the stance foot:
\begin{equation}
\begin{aligned}
C_3:\quad
\Delta z&=\frac{1}{9}\sum_{j=1}^{9}h_j-h_{\mathrm{stance}},\\
|\Delta z|&\le z_{\max}.
\end{aligned}
\label{eq:criterion_reachability}
\end{equation}
A candidate foothold is accepted when the combined condition
\begin{equation}
\bigl(C_1\vee C_2\bigr)\wedge C_3.
\label{eq:foothold_acceptance}
\end{equation}
holds.

Thus, an accepted foothold has a locally level or steadily inclined patch and remains within the reachable elevation range.
\begin{table}[htbp]
  \centering
  \caption{Behavior of the three criteria on each terrain type.}
  \begin{tabular}{lcccl}
    \toprule
    Terrain & $C_1$ & $C_2$ & $C_3$ & Admitted by \\
    \midrule
    Stairs       & Pass & Pass & Pass & $C_1$, level tread \\
    Slope        & Fail & Pass & Pass & $C_2$, steady increment \\
    Rough ground & Pass & Fail & Pass & $C_1$, locally level \\
    \bottomrule
  \end{tabular}
  \label{tab:foothold-criteria}
\end{table}

As Table~\ref{tab:foothold-criteria} shows, $C_1$ retains stair treads and mildly
uneven ground, $C_2$ retains inclined surfaces that $C_1$ would reject merely for
being tilted, and $C_3$ is independent of local shape and therefore removes
cliffs and unreachable platforms rather than rough patches.

To balance computational efficiency and sampling quality during training, we evaluate 12 candidate foothold patches. Among the candidates that satisfy the acceptance condition above, we select the one with the largest height difference from the current foothold, thereby encouraging the robot to practice steps of varying heights.

\subsubsection{Reference Frame Selection and Trajectory Planning}

The reference frame strongly affects the consistency of foothold commands. A target expressed in the base frame moves in world coordinates as the torso moves during a step, which can destabilize training. In contrast, the stance foot remains approximately fixed when there is no slip or rotation. We therefore express each target foothold relative to the stance-foot frame.

\begin{figure}[htbp]
  \centering
  \begin{tikzpicture}[x=0.72cm,y=0.55cm,font=\scriptsize]
    \draw[fill=blue!18, rounded corners=1.5pt]
      (0,0) rectangle (2.4,1.15);
    \draw[fill=orange!25, rounded corners=1.5pt]
      (2.4,0) rectangle (5.6,1.15);
    \draw[fill=green!20, rounded corners=1.5pt]
      (5.6,0) rectangle (8,1.15);
    \node[align=center] at (1.2,0.58)
      {\textbf{I. Prepare}\\Double support};
    \node[align=center] at (4.0,0.58)
      {\textbf{II. Swing}\\Single support};
    \node[align=center] at (6.8,0.58)
      {\textbf{III. Land}\\Double support};

    \draw[->,thick] (0,-0.22) -- (8.25,-0.22);
    \foreach \x/\phase in {0/0,2.4/0.3,5.6/0.7,8/1}
      {
        \draw (\x,-0.16) -- (\x,-0.28);
        \node[below] at (\x,-0.30) {$\varphi=\phase$};
      }

    \node[anchor=east] at (-0.12,-1.15) {Swing};
    \node[blue!60!black] at (1.2,-1.15) {$\bullet$};
    \node[orange!70!black] at (4.0,-1.15) {$\circ$};
    \node[green!50!black] at (6.8,-1.15) {$\bullet$};
    \node[anchor=east] at (-0.12,-1.65) {Stance};
    \draw[very thick,green!50!black] (0,-1.65) -- (8,-1.65);
  \end{tikzpicture}
  \caption{Three stages of one discrete step. Swing-foot contact
  ($\bullet$) and flight ($\circ$) alternate, while the stance foot remains
  in continuous contact.}
  \label{fig:step_phase}
\end{figure}
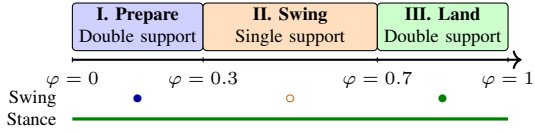

Each step, from command reception to post-touchdown stabilization, consists of three stages: preparation, swing transfer, and landing. During preparation, both feet remain in contact while the swing foot receives its target. It then lifts and moves toward the target while the stance foot supports the robot. After touchdown, both feet remain in contact as the robot stabilizes for the next step. These three stages are illustrated in Fig.~\ref{fig:step_phase}.

For swing-phase height planning, we use a quadratic Bézier profile that combines linear interpolation between the lift-off and landing heights with a parabolic clearance hump. This provides smooth endpoint transitions while maintaining clearance for both upward and downward steps.

Let $s\in[0,1]$ be the normalized swing progress over the transfer window $0.3\leq\varphi\leq 0.7$,
\begin{equation}
  s = \mathrm{clip}\!\left(\frac{\varphi-0.3}{0.4},\,0,\,1\right).
\end{equation}
The reference relative height of the swing foot is
\begin{equation}
  z_{\mathrm{ref}}(s)
  = (1-s)\,z_{\mathrm{start}}
  + s\,z_{\mathrm{target}}
  + 4s(1-s)\,h,
  \label{eq:swing_z_bezier}
\end{equation}
where each term has the following role:
\begin{itemize}
  \item $(1-s)\,z_{\mathrm{start}}+s\,z_{\mathrm{target}}$ is the linear blend between the swing-foot relative height frozen at lift-off, $z_{\mathrm{start}}$, and the terrain-consistent landing height, $z_{\mathrm{target}}$. It interpolates a straight ramp from the current foothold to the next one, so the same planner covers both upward and downward steps.
  \item $4s(1-s)$ is a parabolic (quadratic Bézier) hump that is $0$ at $s=0$ and $s=1$ and reaches $1$ at mid-swing ($s=0.5$). Multiplying by the clearance margin $h$ lifts the foot above the linear ramp during flight and returns it to $z_{\mathrm{target}}$ at landing, reducing terrain collision without changing the endpoints.
\end{itemize}

\begin{figure}[htbp]
  \centering
  \begin{minipage}[b]{0.49\linewidth}
    \centering
    \includegraphics[width=\linewidth]{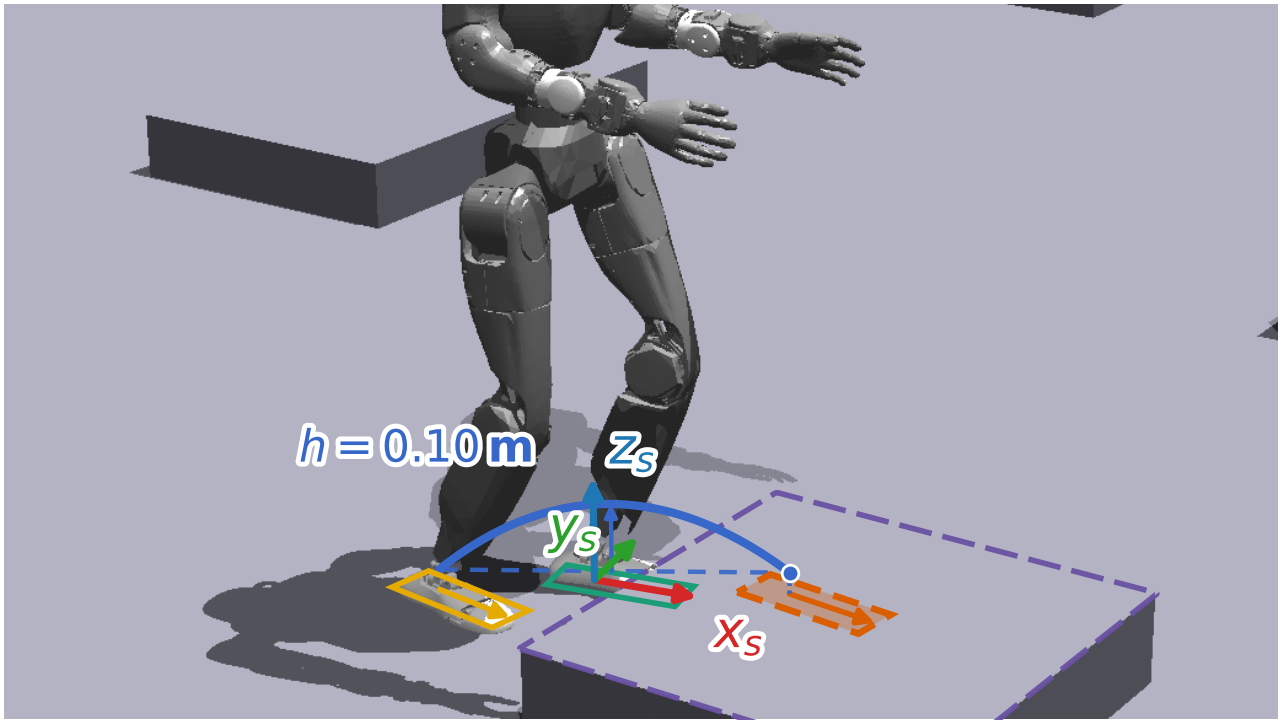}
  \end{minipage}\hfill
  \begin{minipage}[b]{0.49\linewidth}
    \centering
    \includegraphics[width=\linewidth]{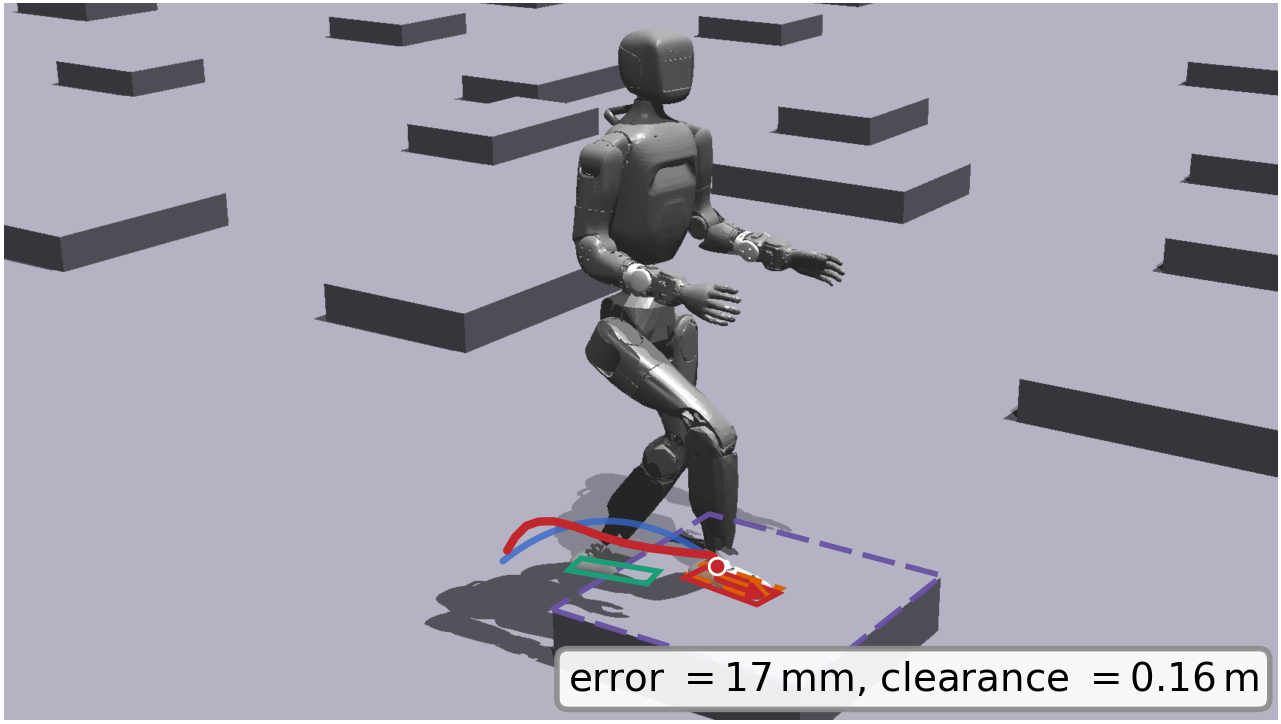}
  \end{minipage}\\[0.2em]
  \includegraphics[width=\linewidth]{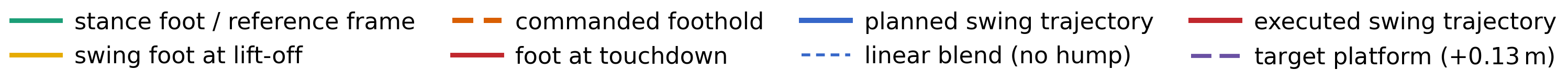}
  \caption{One discrete step-up onto a $0.13\,\mathrm{m}$ platform. (a)~At lift-off (left) the swing foot is commanded $0.37\,\mathrm{m}$ forward and $0.13\,\mathrm{m}$ up, relative to the stance-foot frame, and follows the planned quadratic-B\'ezier $z$ trajectory of~\eqref{eq:swing_z_bezier} with a $0.10\,\mathrm{m}$ clearance hump. (b)~After touchdown (right), the executed swing path follows the planned arc and the foot lands on the commanded foothold.}
  \label{fig:step_traj_sim}
\end{figure}

Fig.~\ref{fig:step_traj_sim} shows the planned swing-foot trajectory in the Isaac Gym simulator and the robot posture after the same step. At each step, one foot is assigned as the stance foot and the other as the swing foot, with the assignment alternating when commands are resampled. Only the swing foot receives the terrain-filtered foothold target and evolving phase; the stance-foot command and phase remain zero.

\begin{figure}[htbp]
  \centering
  \includegraphics[width=0.8\columnwidth]{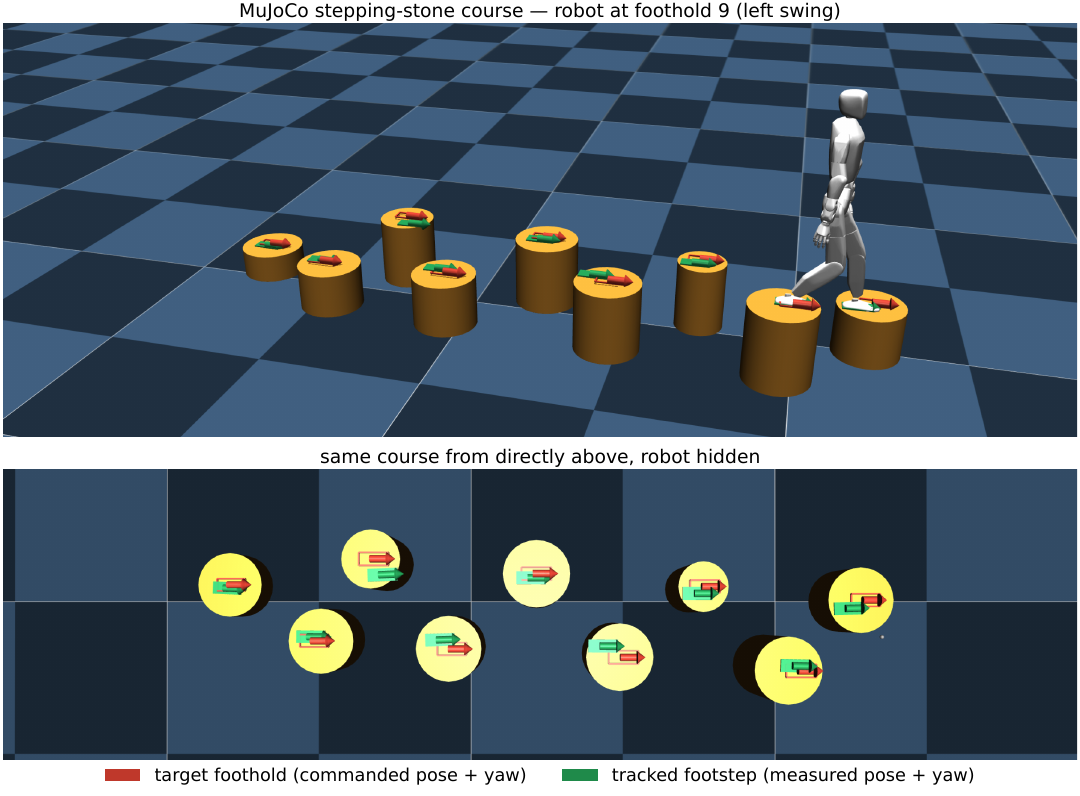}
  \caption{MuJoCo stepping-stone evaluation of LD-PPO foothold tracking.
  Top: the robot traverses sparse cylindrical supports. Bottom: overhead view
  of commanded (red) and executed (green) foothold pose and yaw.}
  \label{fig:foothold_mujoco}
\end{figure}
Fig.~\ref{fig:foothold_mujoco} further illustrates the continuous foothold tracking trajectory on a MuJoCo stepping-stone course, where commanded and executed footholds remain closely aligned in both position and yaw.

\subsection{Decoupled Loco-Manipulation Design}
\subsubsection{Lower-Body Locomotion}
The AMP-based walking teacher shares the foothold teacher's observation history, privileged critic input, and 15-D lower-body action space. This common interface enables subsequent latent distillation. During walking, the command retains the velocity and torso channels while setting the eight foothold channels to zero, as given in~\eqref{eq:walking_command}:
\begin{equation}
\mathbf{c}
=
\bigl[
v_x,\,v_y,\,v_{\mathrm{yaw}},\,
  h,\,r,\,p,\,\psi,\,
\mathbf{0}_{8}
\bigr].
\label{eq:walking_command}
\end{equation}
Commands are periodically resampled, while a curriculum expands the velocity, base-height, and torso-orientation ranges. Training includes both standing and walking environments: walking episodes use an alternating anti-phase gait, whereas standing episodes use zero velocity and fixed leg phases.

To encourage natural motion, AMP matches consecutive transitions of a 30-D lower-body feature---leg joint positions and velocities together with base-frame foot positions---to a reference walking clip. Excluding the arms and waist prevents interference with upper-body disturbances, while a sagittal-plane symmetry loss reduces left--right asymmetry.

\subsubsection{Upper-Body Manipulation}

The upper-body policy tracks torso-relative hand poses generated by retargeting AMASS motions through forward kinematics (FK). Its observation comprises the target position and orientation of each hand, the positions and velocities of 14 arm joints, and the previous action. The actor input stacks the current observation with five preceding frames. The critic additionally receives privileged FK joint-reference residuals and head and torso contact states, while the actor relies exclusively on deployable Cartesian commands and proprioceptive measurements.

\subsection{Encoder--Decoder-Based LD-PPO}

\begin{figure}[t]
  \centering
  \includegraphics[width=0.48\textwidth,height=0.26\textheight,keepaspectratio]{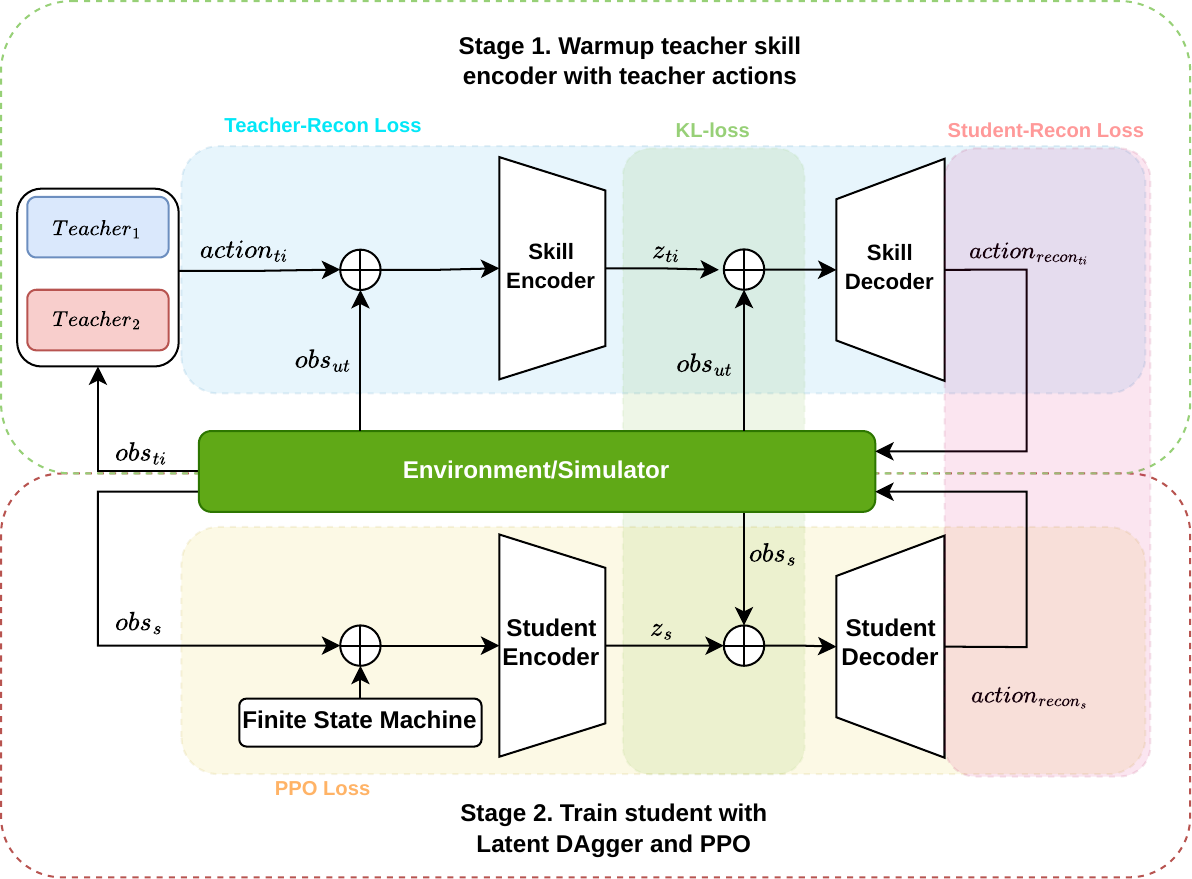}
  \caption{LD-PPO training framework. Top: the mode-selected expert
  action and observation are used to train the teacher skill autoencoder during warm-up,
  producing the target latent representation $\mathbf{z}^{T}_t$. Bottom: the
  student collects on-policy states in the simulator and learns through PPO,
  expert-action reconstruction, and KL alignment between
  $\mathbf{z}^{S}_t$ and $\mathbf{z}^{T}_t$. Only the student encoder--decoder
  path is required at deployment.}
  \label{fig:fig2}
\end{figure}

After pretraining the foothold-tracking and locomotion policies, we fuse them into a single student that supports both continuous walking and discrete stepping. We use distillation-PPO (D-PPO) as the baseline. Although vanilla D-PPO can transfer the basic expert behaviors, we identify two limitations for this multi-gait setting. First, vanilla D-PPO supervises the student mainly through pointwise joint-space action regression, as in~\eqref{eq:action_distillation}:
\begin{equation}
\mathcal{L}_{\mathrm{act}}
=
\left\|
\boldsymbol{\mu}_{S}(\mathbf{o}_t)
-
\mathbf{a}^{T}_t
\right\|_2^2.
\label{eq:action_distillation}
\end{equation}
Although the student shares features across joints, this element-wise objective does not explicitly preserve the cross-joint correlations that generate coordinated expert actions. This matters for stepping, where swing-leg motion, stance-foot stability, contact timing, and torso compensation must remain synchronized. Small joint-space errors can therefore accumulate into Cartesian foothold error, foot slip, or loss of balance. Action-level imitation also provides no direct constraint on a shared representation across substantially different gait and contact modes.

Second, direct regression requires the student to discover the coordination structure implicitly while mapping a high-dimensional observation history to teacher actions. This can be inefficient when standing, walking, and stepping require substantially different control strategies. We therefore augment D-PPO with teacher-conditioned latent alignment, providing complementary supervision in action space and skill-representation space.

At each time step, the state machine determines the active mode $m_t\in\{\mathrm{stand},\mathrm{walk},\mathrm{step}\}$ from the command and phase variables contained in the observation history. The corresponding frozen expert supplies the DAgger target,
\begin{equation}
\mathbf{a}^{T}_t=\pi^{T}_{m_t}(\mathbf{o}_t).
\label{eq:teacher_action}
\end{equation}
The student replaces the conventional MLP actor with an encoder--decoder policy. Its encoder takes only deployable observations and predicts a Gaussian latent posterior, while its decoder conditions on both the observation and latent code, as defined in~\eqref{eq:student_encoder_decoder}:
\begin{equation}
\begin{aligned}
(\boldsymbol{\mu}^{S}_t,\log\boldsymbol{\sigma}^{S\,2}_t)
  &=E_S(\mathbf{o}_t),\\
\boldsymbol{\sigma}^{S}_t
  &=\exp\bigl(\tfrac12\log\boldsymbol{\sigma}^{S\,2}_t\bigr),\\
\mathbf{z}^{S}_t
  &=\boldsymbol{\mu}^{S}_t+\boldsymbol{\sigma}^{S}_t\odot\boldsymbol{\epsilon}^{S}_t,\\
\hat{\mathbf{a}}^{S}_t
  &=D_S([\mathbf{o}_t,\mathbf{z}^{S}_t]),
\end{aligned}
\label{eq:student_encoder_decoder}
\end{equation}
where \(\boldsymbol{\epsilon}^{S}_t\sim\mathcal{N}(\mathbf{0},\mathbf{I})\).
The observation contains the state-machine command and phase variables, so the student can infer the active mode at deployment without access to the teacher action. The latent is therefore not a mode label; it represents the action-producing coordination context within that mode.

To construct the target skill-representation space, we introduce a teacher skill encoder--decoder with the same latent dimensionality as the student. In contrast to the student encoder, the teacher encoder observes both the state and the selected teacher action:
\begin{equation}
\begin{aligned}
(\boldsymbol{\mu}^{T}_t,\log\boldsymbol{\sigma}^{T\,2}_t)
  &=E_T([\mathbf{o}_t,\mathbf{a}^{T}_t]),\\
\boldsymbol{\sigma}^{T}_t
  &=\exp\bigl(\tfrac12\log\boldsymbol{\sigma}^{T\,2}_t\bigr),\\
\mathbf{z}^{T}_t
  &=\boldsymbol{\mu}^{T}_t+\boldsymbol{\sigma}^{T}_t\odot\boldsymbol{\epsilon}^{T}_t,\\
\hat{\mathbf{a}}^{T}_t
  &=D_T([\mathbf{o}_t,\mathbf{z}^{T}_t]),
\end{aligned}
\label{eq:teacher_encoder_decoder}
\end{equation}
where \(\boldsymbol{\epsilon}^{T}_t\sim\mathcal{N}(\mathbf{0},\mathbf{I})\).
Conditioning the teacher encoder on the state--action pair allows its latent to encode action-producing coordination that is not explicitly available to the student at deployment, such as the balance and swing strategy selected by the expert. The teacher decoder reconstructs the expert action, thereby defining a task-relevant latent space for the heterogeneous expert behaviors. The student is then trained to infer a corresponding latent using deployable observations only.

The action-space reconstruction objective is given by~\eqref{eq:reconstruction}:
\begin{equation}
\mathcal{L}_{\mathrm{rec}}
=\frac{1}{2}\left(
\|\hat{\mathbf{a}}^{T}_t-\mathbf{a}^{T}_t\|_2^2+
\|\hat{\mathbf{a}}^{S}_t-\mathbf{a}^{T}_t\|_2^2
\right).
\label{eq:reconstruction}
\end{equation}
We further align the student posterior with the teacher skill posterior through~\eqref{eq:latent_alignment}:
\begin{equation}
\mathcal{L}_{\mathrm{align}}
=D_{\mathrm{KL}}\!\left(
q_S(\mathbf{z}_t\mid\mathbf{o}_t)
\;\|\;
q_T(\mathbf{z}_t\mid\mathbf{o}_t,\mathbf{a}^{T}_t)
\right),
\label{eq:latent_alignment}
\end{equation}
and define the LD-PPO objective in~\eqref{eq:ld_ppo}:
\begin{equation}
\mathcal{L}_{\mathrm{LD}}
=\mathcal{L}_{\mathrm{rec}}+\beta\mathcal{L}_{\mathrm{align}},\qquad
\mathcal{L}
=(1-\alpha)\mathcal{L}_{\mathrm{PPO}}+\alpha\mathcal{L}_{\mathrm{LD}}.
\label{eq:ld_ppo}
\end{equation}
The reconstruction terms preserve action-level fidelity, while posterior alignment encourages the student to use the same task-relevant coordination representation as the teacher. The weight \(\beta\) balances the two imitation signals and keeps action reconstruction as the primary objective. During the initial warm-up, only the teacher autoencoder is trained with the reconstruction objective, establishing the latent target before student alignment begins. The student then learns to map deployable observations to the latent coordination representation and decode it into expert-like joint actions.
\section{Experiments}

\begin{table*}[htbp]
  \centering
  \begin{threeparttable}
  \footnotesize
  \setlength{\tabcolsep}{12.1pt}
  \caption{Real-robot ablation of the four LD-PPO ingredients}
  \label{tab:ablation}
  \begin{tabular}{lcccc rr rrrr}
    \toprule
    & \multicolumn{4}{c}{Ingredient}
    & \multicolumn{2}{c}{Foothold tracking}
    & \multicolumn{4}{c}{Body-posture tracking} \\
    \cmidrule(lr){2-5}\cmidrule(lr){6-7}\cmidrule(lr){8-11}
    Variant & E--D & Lat. & DAg. & PPO
      & $e_{\mathrm{2D}}$ & $e_{\mathrm{3D}}$
      & Height & Roll & Pitch & Yaw \\
    &&&& & (mm) & (mm) & (cm) & (deg) & (deg) & (deg) \\
    \midrule
    LD-PPO (ours) & \cmark & \cmark & \cmark & \cmark
      & $\mathbf{113.2}$ & $\mathbf{136.6}$
      & $\mathbf{3.428}$ & $\mathbf{0.998}$ & $\mathbf{0.964}$ & $4.582$ \\
    w/o latent & \cmark & \xmark & \cmark & \cmark
      & $144.1$ & $169.0$
      & $4.909$ & $1.980$ & $1.697$ & $\mathbf{3.930}$ \\
    w/o enc--dec & \xmark & \xmark & \cmark & \cmark
      & $125.8$ & $150.5$
      & $3.846$ & $1.454$ & $2.010$ & $4.191$ \\
    w/o DAgger & \cmark & \cmark & \xmark & \cmark
      & $445.9$ & $467.3$
      & $7.566$ & $1.437$ & $2.231$ & $4.322$ \\
    w/o PPO & \cmark & \cmark & \cmark & \xmark
      & $160.3$ & $160.3$
      & $3.766$ & $1.302$ & $1.567$ & $4.460$ \\
    \bottomrule
  \end{tabular}
  \begin{tablenotes}
    \item[] E--D denotes the encoder--decoder actor, Lat.\ teacher-conditioned
      latent alignment, and DAg.\ DAgger expert-action supervision. Lower
      errors are better; bold marks the best result in each metric.
  \end{tablenotes}
  \end{threeparttable}
\end{table*}

For training, we use a three-layer MLP for the two teachers and an encoder--decoder structure for the skill and student policies. Each policy runs at $50\,\mathrm{Hz}$ in the Isaac Gym simulator. Training is conducted on a single RTX 4090 GPU with 4096 environments. 

For the real-robot experiments, we deploy the trained policies on the TianGong
Omni platform, where a MetaCam EDU sensor reconstructs the terrain map at
$10\,\mathrm{Hz}$ and the Manifold Odin1 provides global localization, both
running on an NVIDIA Jetson Thor. A command generator converts localization
feedback and specified footholds through forward kinematics into commanded
foothold pose and yaw. Motion clips provide upper-body targets to
\(\pi_{\mathrm{upper}}\) and posture commands to \(\pi_{\mathrm{student}}\).
On hardware, we evaluate gait switching with ground-level manipulation, stair
climbing while carrying a box, and foothold tracking on stepping stones of
different heights. Each experiment is repeated 20 times, and the collected
measurements are aggregated for evaluation.

\subsection{Distillation Performance}

We evaluate the four components of LD-PPO: the encoder--decoder actor,
teacher-conditioned latent alignment, DAgger action supervision, and on-policy
PPO refinement, through structured ablations under identical commands:
\begin{itemize}
  \item \textbf{LD-PPO (ours)}: encoder--decoder actor with all losses,\\
  \(\mathcal{L}=(1-\alpha)\mathcal{L}_{\mathrm{PPO}}+\alpha(\mathcal{L}_{\mathrm{rec}}+\beta\mathcal{L}_{\mathrm{align}})\).
  \item \textbf{w/o latent} (encoder--decoder D-PPO): \\ \(\beta=0\),
  \(\mathcal{L}=(1-\alpha)\mathcal{L}_{\mathrm{PPO}}+\alpha\mathcal{L}_{\mathrm{rec}}\).
  \item \textbf{w/o enc--dec} (vanilla D-PPO): \\ MLP actor and \(\beta=0\),
  \(\mathcal{L}=(1-\alpha)\mathcal{L}_{\mathrm{PPO}}+\alpha\mathcal{L}_{\mathrm{act}}\).
  \item \textbf{w/o DAgger}: student action reconstruction is removed,
  \(\mathcal{L}=(1-\alpha)\mathcal{L}_{\mathrm{PPO}}+\alpha\beta\mathcal{L}_{\mathrm{align}}\).
  \item w/o PPO: \(\alpha=1\),
  \(\mathcal{L}=\mathcal{L}_{\mathrm{rec}}+\beta\mathcal{L}_{\mathrm{align}}\).
\end{itemize}
All five variants are evaluated on the real
robot, while their learning dynamics are compared in simulation. At the first
touchdown of each commanded step, we measure horizontal error
$e_{\mathrm{2D}}=\sqrt{e_x^2+e_y^2}$ and spatial error
$e_{\mathrm{3D}}=\sqrt{e_x^2+e_y^2+e_z^2}$ in the stance-foot command frame.
We also report mean absolute errors in body height and orientation.

As shown in Table~\ref{tab:ablation}, LD-PPO achieves the lowest real-robot
foothold errors, with $e_{\mathrm{2D}}=113.2\,\mathrm{mm}$ and
$e_{\mathrm{3D}}=136.6\,\mathrm{mm}$. Relative to vanilla D-PPO, these values
represent reductions of $10.0\%$ and $9.2\%$, respectively. Removing latent
alignment increases the errors to $144.1$ and $169.0\,\mathrm{mm}$, indicating
that the shared latent representation improves transfer beyond architectural
factorization alone. DAgger has the largest effect: without direct expert-action reconstruction,
$e_{\mathrm{3D}}$ rises to $467.3\,\mathrm{mm}$ even though
teacher-conditioned latent alignment remains active. Removing PPO also
degrades foothold accuracy despite retaining the two distillation losses. The
complete method yields the lowest height, roll, and pitch errors, whereas the
variant without latent alignment has the lowest yaw error. Notably, removing
DAgger substantially increases foothold errors while leaving orientation
errors comparatively moderate. This disparity suggests that posture tracking
is easier to learn and transfer than precise foothold tracking, making
foothold accuracy a more discriminative measure of distillation quality.

\begin{figure}[htbp]
  \centering
  \includegraphics[width=\columnwidth]{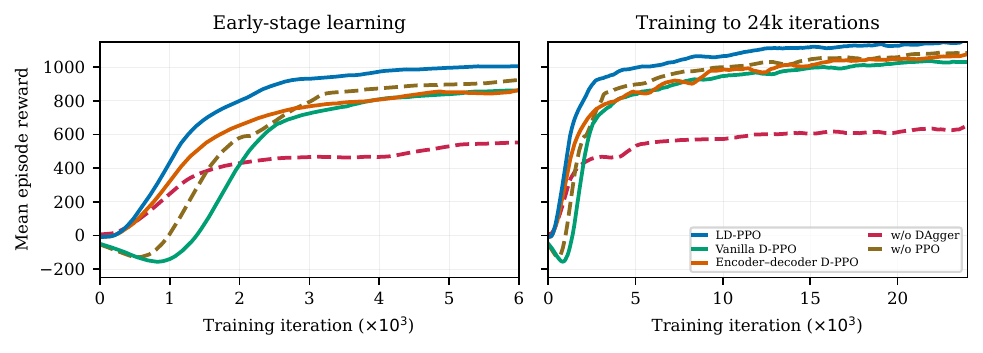}
  \caption{Simulation training curves for LD-PPO, the two D-PPO comparisons,
  and the variants without DAgger or PPO (dashed). The left panel expands the
  first $6\,000$ iterations; both panels use the same vertical scale.}
  \label{fig:distillation_training}
\end{figure}

The training curves in Fig.~\ref{fig:distillation_training} complement the
tracking results. LD-PPO and encoder--decoder D-PPO improve faster initially
than vanilla D-PPO, indicating that the structured actor facilitates
distillation. Although encoder--decoder D-PPO has larger final foothold errors
than vanilla D-PPO in Table~\ref{tab:ablation}, its faster reward improvement
is a practical optimization advantage; the full LD-PPO results show that
latent alignment is required to translate this advantage into higher final
tracking accuracy. Without DAgger, the reward plateaus at a substantially lower value even though latent alignment is retained; without PPO, training reward remains high
although foothold error increases to $160.3\,\mathrm{mm}$. Consequently,
training reward alone does not establish accurate task execution. Together,
the results show that DAgger provides action-level guidance, latent alignment
improves multi-gait transfer, and PPO refines the policy on the
student's state distribution.

\subsection{Manipulation Performance}

We evaluate the Cartesian upper-body policy in simulation and on the real robot,
measuring wrist position error
as Euclidean distance and orientation error as the geodesic angle between
commanded and measured quaternions. The real-robot entries pool three trials of
the complete LD-PPO system.

\begin{table}[htbp]
  \centering
  \scriptsize
  \setlength{\tabcolsep}{1pt}
  \caption{Upper-policy end-effector tracking performance}
  \label{tab:manipulation_tracking}
  \begin{tabular}{llrrrrr}
    \toprule
    Domain & Wrist & $n$ & Pos. MAE (cm) & Median (cm) & P95 (cm)
          & Ori. MAE (deg) \\
    \midrule
    Simulation & Left  & 22,000 & 2.67 & 1.73 & 10.32 &  7.92 \\
               & Right & 22,000 & 2.46 & 1.79 &  7.92 &  8.60 \\
    \hdashline
    Real robot & Left  & 10,000 & 2.07 & 2.22 & 2.31 & 12.70 \\
               & Right & 10,000 & 2.23 & 2.38 & 2.43 & 13.90 \\
    \bottomrule
  \end{tabular}
\end{table}

As shown in Table~\ref{tab:manipulation_tracking}, both wrists achieve position MAE below $2.7\,\mathrm{cm}$ in simulation and below $2.3\,\mathrm{cm}$ on the real robot, with orientation MAE below $14^\circ$.

\subsection{Analysis and Results}

The experimental results show that the encoder--decoder structure accelerates
early distillation, although it does not independently reduce final foothold
error. Combined with teacher-conditioned latent alignment, however, it enables
LD-PPO to achieve the lowest 2D and 3D errors among all evaluated variants.
DAgger action supervision has the largest effect on foothold accuracy, while
PPO refinement further improves tracking on the student's state distribution.
The posture results also show that accurate quasi-static tracking does not by
itself imply accurate foothold placement.

The manipulation experiments further show that the upper-body policy maintains accurate end-effector tracking in both simulation and real-robot experiments.

\section{Conclusion}

We presented a hierarchical framework for 3D locomotion and manipulation that
combines foothold tracking, AMP-based velocity tracking, and end-effector
control. Target sampling and swing-trajectory planning facilitate precise
foothold placement, while LD-PPO transfers the teachers' complementary
skills into one student policy. Simulation and real-robot experiments show
that the resulting controller performs accurate whole-body locomotion and
manipulation.

\section*{ACKNOWLEDGMENT}
The authors thank Liguo Zhang, Shichao Liang, and Guangyuan Yu
of the Humanoid Innovation Department at X-Humanoid for
repairing and maintaining the TianGong Omni platform and for
providing hardware support during the real-robot experiments.

\bibliographystyle{IEEEtran}
\bibliography{references}

\end{document}